\documentclass[letterpaper,10pt,conference]{ieeeconf}

\IEEEoverridecommandlockouts
\usepackage{graphicx}
\usepackage{amsmath}
\usepackage{amssymb}
\usepackage{booktabs}
\usepackage{cite}
\usepackage{url}
\usepackage{stfloats}

\title{\LARGE \bf
TS-WAM: Temporal Latent Actions and World-Action Coupling for
Vision-Language Robot Control
}

\author{\smash{\begin{tabular}[t]{@{}c@{}}
\normalsize Menghao Yin, Jiaju Yin, Xiyue Dong, Lixin Xu,
Zhenhui Zhang, Ke Zhang, Yating Feng, Renjing Xu \\[-0.5ex]
\normalsize The Hong Kong University of Science and Technology (Guangzhou)
\\
\end{tabular}}}

\begin{document}

\maketitle
\thispagestyle{empty}
\pagestyle{empty}

\begin{abstract}
Many vision-language-action policies rely on action-labeled robot
demonstrations and lack an explicit mechanism for predicting future visual
states. These limitations hinder learning from action-free videos and can
encourage overfitting to demonstration conditions, weakening generalization
and robustness. We present TS-WAM, a lightweight Teacher-Student World-Action
Model that addresses these challenges through self-supervised temporal
learning and explicit future prediction. Its vlm-jepa framework learns from
heterogeneous data, including robot demonstrations and action-free human
videos, using an EMA target branch without requiring a separately trained
external teacher. By coupling vlm-jepa with delta-lwm, TS-WAM learns to
predict latent actions from the current observation and instruction and uses
them to predict future visual features that guide action generation at
inference. A unified end-to-end training architecture jointly optimizes
temporal representation learning, visual dynamics, and robot control without
a separate latent-action or world-model pretraining stage. The EMA target
branch is removed at inference, keeping the deployed policy lightweight.
Experiments on LIBERO, LIBERO-Plus, and SimplerEnv, together with real-world
manipulation experiments on Franka Research 3 and RevoTron, demonstrate
that TS-WAM achieves strong manipulation performance, generalization, and
robustness.
\end{abstract}

\section{INTRODUCTION}
\label{sec:introduction}

A robot must convert a language instruction and a changing visual scene into a
temporally coherent sequence of actions. Recent vision-language-action (VLA)
systems use a pretrained vision-language model (VLM) as a semantic backbone
and attach a diffusion or flow-matching action decoder
\cite{chi2024diffusion,kim2024openvla,starvla2025,xvla2025}.
This recipe gives strong instruction following, but
the action interface is commonly learned directly from robot trajectories. The
resulting hidden state need not represent the short-horizon visual change that
an action is intended to cause. Consequently, the policy has no explicit
internal prediction of what its action will do before it is executed.

At the same time, large video collections contain abundant information about
how objects and hands move, but generally do not provide robot actions. Latent
action models (LAMs) address this gap by compressing temporal changes into a
compact code \cite{ye2025lapa}. Existing systems, however, often pretrain a
LAM or a latent world model separately and subsequently adapt it to a robot
policy. This multi-stage procedure introduces an additional checkpoint and
optimization boundary, and it does not guarantee that the learned latent is
useful to the final action decoder.

We propose TS-WAM (Teacher-Student World-Action Model), a task-adaptation
architecture that makes temporal prediction the central interface of a VLA
policy without requiring a separately pretrained latent-action model. The
model uses an asymmetric pair of VLM branches.

\begin{figure}[!t]
    \centering
    \includegraphics[width=\columnwidth]{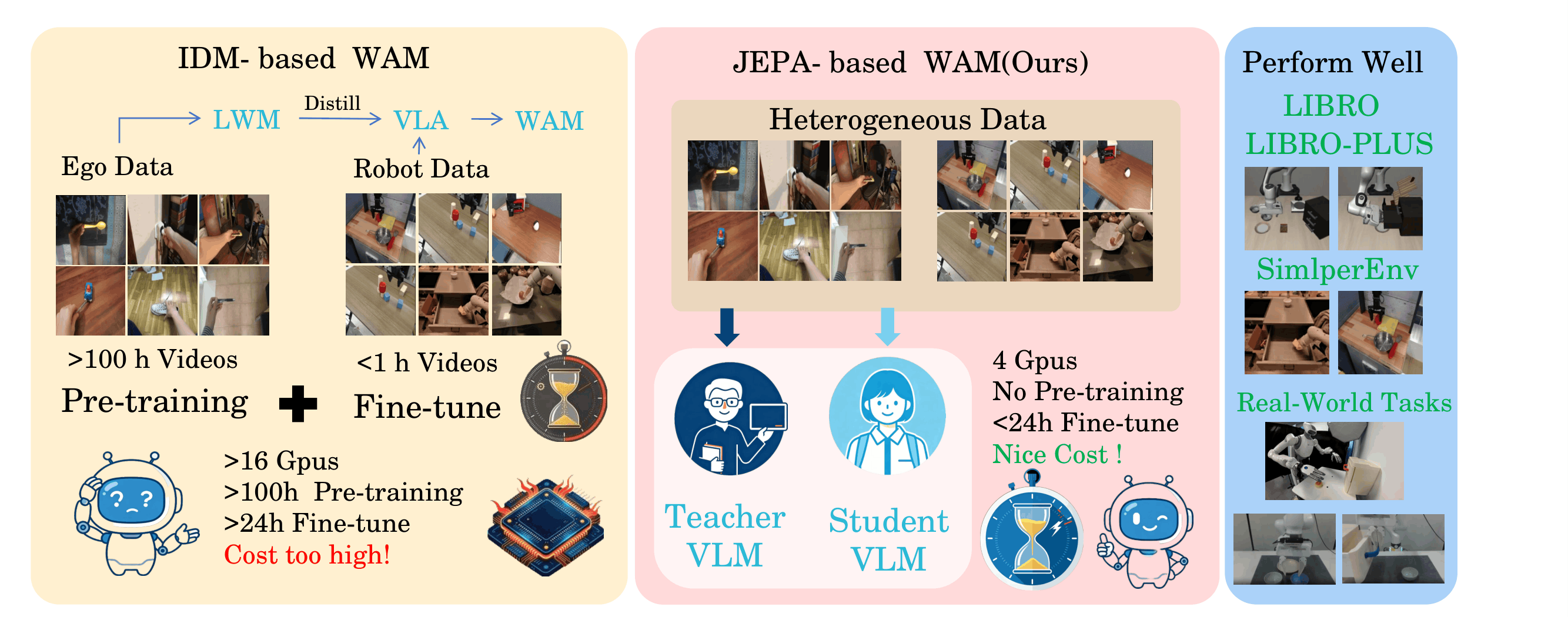}
    \caption{\textbf{Comparison of conventional IDM-based world-action models and
    our JEPA-based TS-WAM.} Pre-training followed by fine-tuning often requires
    substantial computational resources and time, whereas TS-WAM achieves
    strong performance with limited resources through end-to-end training
    without a separate pre-training stage.}
    \label{fig:teaser}
\end{figure}

The online $X$-VLM
receives only the current observation and language instruction. It produces
latent-$X$ tokens that a small transformer predictor maps to a predicted latent
action. The EMA $Y$-VLM receives the current and future observations and reads
the corresponding target latent action. This temporal information asymmetry
turns future visual evidence into a training-only teacher signal while keeping
the deployed policy causal.

The latent action is coupled to visual dynamics and control. A frozen DINO
encoder represents the current and future images as spatial feature grids. A
newly initialized latent world decoder predicts the feature displacement
$\Delta F_t=F_{t+1}-F_t$ conditioned on the current feature and latent action.
The predicted future feature $F_t+\widehat{\Delta F}_t$ is passed to a
conditional flow-matching action head together with VLM context and robot
state. The policy therefore follows an explicit chain: infer a latent intent,
predict its visual consequence, and ground that consequence into an action
chunk.

The same training procedure can mix robot demonstrations and action-free human
videos. Human clips contribute current--future visual pairs to the teacher and
world-model objectives, whereas robot clips additionally provide state and
action supervision for the flow head. Thus, action-free data expands temporal
supervision without providing embodiment-specific action labels. The
task-specific world decoder is initialized from scratch and co-adapts with the
VLM and policy.

This joint objective keeps temporal prediction aligned with the downstream
control interface throughout task adaptation. At deployment, the resulting
policy uses only the current observation while retaining visual anticipation
learned from both data sources.

Our main contributions are as follows:
\begin{itemize}
    \item We introduce vlm-jepa, a self-supervised framework for learning
    temporal latent-action representations from robot demonstrations and
    action-free human videos.
    \item We propose TS-WAM, a lightweight world-action model that unifies
    temporal representation learning, visual prediction, and action generation
    through end-to-end training. It anticipates action consequences to guide
    control, without separate latent-action or world-model pretraining.
    \item Experiments on LIBERO, LIBERO-Plus, and SimplerEnv, together with
    real-world evaluations on Franka Research 3 and RevoTron, demonstrate
    TS-WAM's strong manipulation performance.
\end{itemize}

\begin{figure*}[t]
    \centering
    \includegraphics[width=\textwidth]{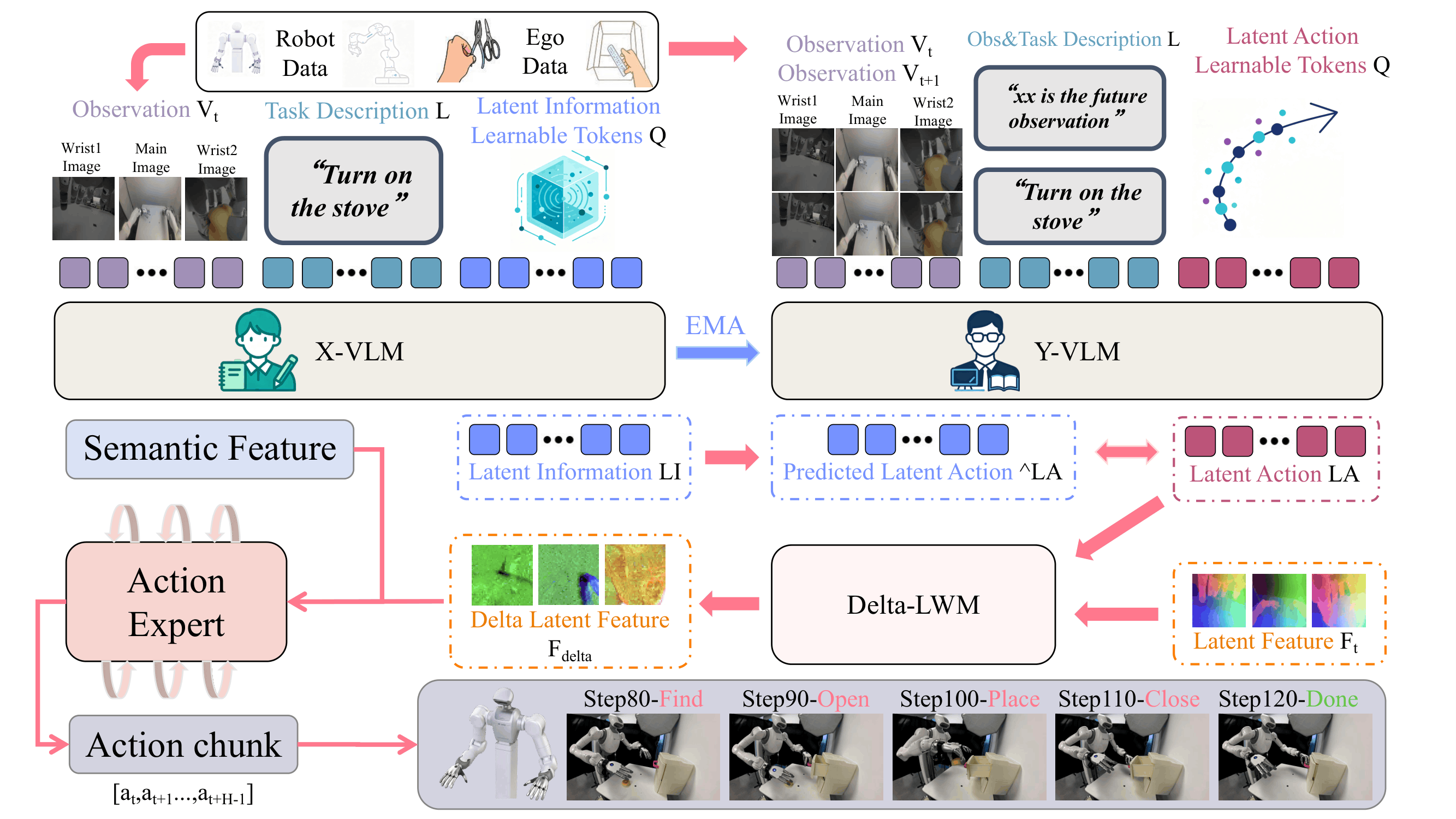}
    \caption{\textbf{Overview of TS-WAM.} Vlm-jepa predicts temporal latent actions,
    delta-lwm predicts visual feature changes, and the flow-matching action
    expert generates robot actions conditioned on predicted future features
    and VLM context. The EMA teacher provides supervision during training.}
    \label{fig:overview}
\end{figure*}

\section{RELATED WORK}
\label{sec:related_work}

\textbf{Vision-language-action policies.}
VLA methods adapt large multimodal models to robot demonstrations and decode
continuous action chunks with autoregressive, diffusion, or flow-based heads
\cite{kim2024openvla,starvla2025,xvla2025}.
Their pretrained semantic representations are useful for language grounding,
as demonstrated by recent generalist policy and multimodal-control systems
\cite{wang2025univla,maes2026leworldmodel,parthasarathy2025closing},
but the action decoder is typically the only component trained to model robot
dynamics. TS-WAM retains the VLM as a semantic context encoder while adding an
explicit temporal latent-action and future-feature interface.

\textbf{Latent actions and world models.}
World models learn compact dynamics for planning and control
\cite{ha2018worldmodels,hafner2019planet,chua2018pets,hafner2019dream,
hafner2023dreamerv3,maes2026leworldmodel,sobal2025pldm}.
Reconstruction-free models such as DINO-WM predict features from a frozen
visual representation \cite{zhou2024dino,oquab2024dinov2}, avoiding the cost of pixel
prediction. Existing approaches commonly first pretrain on large-scale video
or interaction data to obtain latent-action or dynamics supervision, and then
fine-tune on robot demonstrations for embodiment-specific control
\cite{ye2025lapa,bu2025univla}. This two-stage pipeline incurs substantial
additional computational cost and resource demands. TS-WAM instead
jointly optimizes a causally predicted latent action and world-action model
within the deployed VLM.

\textbf{Joint-embedding predictive architectures.}
JEPA methods learn by predicting the representation of a target region or
future observation from an encoded context, rather than reconstructing the
input in pixel space. A context encoder and predictor estimate the target
embedding, which is provided by a separately encoded target. The target
encoder is commonly updated by an exponential moving average, providing a
stable regression target without explicit negative samples. This objective
emphasizes predictable semantic and temporal structure. I-JEPA and V-JEPA
apply this principle to images and videos, respectively
\cite{assran2023ijepa,bardes2024vjepa,assran2025vjepa2,zhang2026deltajepa}.
Their EMA target encoders provide stable self-supervision for visual
representation learning.
TS-WAM transfers this principle to VLA: the target is a latent action extracted
from a future-aware VLM, while the student is restricted to current
observations, making the learned representation directly usable for control.

\textbf{Learning from heterogeneous video.}
Self-supervised video learning extracts temporal representations without action
labels \cite{bardes2022vicreg,bardes2024vjepa}, but they must still be grounded
in embodiment-specific robot actions. LAPA and UniVLA address this gap by
learning latent actions from videos and adapting them to robot control
\cite{ye2025lapa,bu2025univla}. TS-WAM uses unlabeled human videos for temporal
supervision while jointly optimizing robot actions. Its shared latent-action and
future-feature interfaces align video-derived representations with the robot
objective throughout adaptation.

A second challenge is retaining video-derived knowledge during robot adaptation.
Sequential objectives can cause catastrophic forgetting
\cite{kirkpatrick2017overcoming}, motivating continued video supervision during
robot adaptation. TS-WAM retains
ego-video supervision while learning the robot action objective, anchoring the
shared temporal representation to both visual domains. This joint training is
intended to mitigate forgetting; verifying skill retention requires dedicated
evaluation.

\section{Method}
\label{sec:method}

TS-WAM introduces a novel integration of the JEPA architecture and delta-lwm
with an end-to-end joint training strategy that learns manipulation skills
directly from heterogeneous datasets of robot demonstrations and ego videos.
Figure~\ref{fig:overview} presents the overall architecture.

\subsection*{\textbf{A. Vlm-jepa: temporal latent-action learning}}
\label{sec:jepa}

\begin{figure}[t]
    \centering
    \includegraphics[width=\columnwidth]{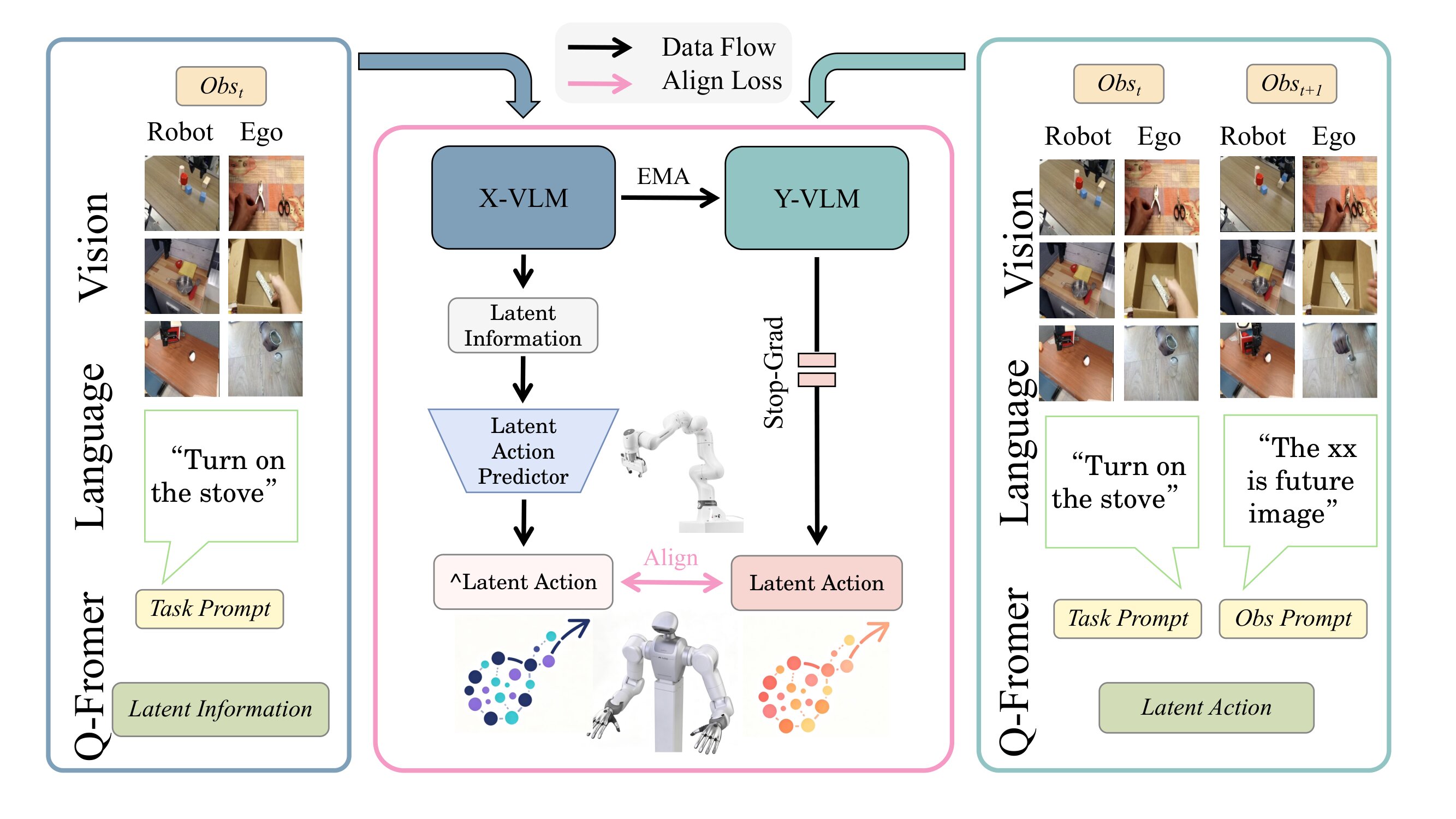}
    \caption{\textbf{Vlm-jepa architecture.} The online $X$-VLM and predictor infer
    latent actions from the current observation and instruction; the EMA
    $Y$-VLM uses the future frame to provide a stop-gradient alignment target.}
    \label{fig:vlm_jepa}
\end{figure}

Fig.~\ref{fig:vlm_jepa} details the future-distilled latent-action interface,
our first contribution. The asymmetric observation inputs allow future-aware
temporal supervision while keeping the online prediction causal.

\noindent\textbf{Online latent-action prediction.}
The online branch inserts $Q_a$ learned latent-$X$ query tokens into the
current image-language sequence. The resulting hidden states are
$X_t\in\mathbb{R}^{Q_a\times d}$. A lightweight transformer predictor
$P_\phi$ maps them to the latent-action space:
\begin{equation}
    \widehat z_t=P_\phi(X_t,Q_Y),
    \qquad \widehat z_t\in\mathbb{R}^{Q_a\times d_z},
    \label{eq:student}
\end{equation}
where $Q_Y$ are learned target-query embeddings. The online branch has access
only to information available before the action is executed.

More explicitly, the predictor projects the latent-$X$ tokens and target
queries into a shared space, adds separate learned positional embeddings, and
applies a self-attention stack to their concatenation. Only the final target
positions are retained and projected back to the world-model code space:
\begin{equation}
 \widehat z_t=W_o\operatorname{LN}\!\left(
 \operatorname{Tr}_\phi\!\left[W_xX_t+E_x\ ;\ W_qQ_Y+E_y\right]_{\rm target}
 \right).
 \label{eq:predictor_detail}
\end{equation}
This explicit bottleneck prevents the latent action from being copied directly
from a single VLM hidden state and forces temporal intent to be inferred from
the current observation.

\noindent\textbf{Future-aware EMA teacher.}
The teacher branch uses the same VLM and latent-action mapper but receives the
current and future images. It reads target query tokens and produces target
$z_t^Y$. Teacher parameters are updated after each optimizer step by
\begin{equation}
    \xi\leftarrow m\xi+(1-m)\theta_Y,
    \label{eq:ema}
\end{equation}
where $m$ is the EMA momentum. Gradients do not flow through the teacher. We
optionally normalize the detached target over its code dimension and align the
online prediction with it using
\begin{equation}
    \bar z_t^Y=\operatorname{LN}_{z}(\operatorname{sg}(z_t^Y)),
    \qquad
    \mathcal{L}_{\mathrm{temp}}
    =\left\|\widehat z_t-\bar z_t^Y\right\|_1.
    \label{eq:temp}
\end{equation}
This objective does not reconstruct pixels or require an action label. It asks
the student to infer the temporal change represented by the privileged teacher
from the current observation alone.

At inference, the teacher is pruned. The online VLM and predictor therefore
implement a causal map $(o_t,\ell)\mapsto\widehat z_t$ with no extra temporal
input or inference-time teacher cost.

\subsection*{\textbf{B. Delta-lwm: latent world-action coupling}}
\label{sec:world_action}

\noindent\textbf{One-step feature-delta visual dynamics.}
We use a frozen DINOv3 encoder to obtain spatial feature grids $F_t$ and
$F_{t+1}$ from the current and future frames. Instead of predicting an
absolute future representation, the trainable decoder $D_\psi$ predicts the
motion target
\begin{equation}
    \Delta F_t=F_{t+1}-F_t,
    \qquad
    \widehat{\Delta F}_t=D_\psi(F_t,\widehat z_t).
    \label{eq:delta}
\end{equation}
The decoder is initialized from scratch for every TS-WAM run. This is a
one-step visual dynamics model rather than a complete long-horizon simulator.
During training,
both the teacher and student latent paths supervise the same target:
\begin{equation}
\mathcal{L}_{\mathrm{world}}=
\left\|D_\psi(F_t,z_t^Y)-\Delta F_t\right\|_2^2+
\left\|D_\psi(F_t,\widehat z_t)-\Delta F_t\right\|_2^2.
\label{eq:world}
\end{equation}
The first term supplies a stable target-side route; the second trains the
latent available to the deployed policy. Predicting a feature displacement
focuses the decoder on action-relevant visual change and makes the predicted
future feature explicit:
\begin{equation}
    \widehat F_{t+1}=F_t+\widehat{\Delta F}_t.
    \label{eq:future}
\end{equation}

The decoder is an AdaLN-conditioned transformer over the spatial tokens of
$F_t$. Its conditioning vector is the pooled latent action after a learned
projection, so the same visual state can produce different displacements for
different inferred intents. To make this dependence identifiable rather than
incidental, we compare the matched latent with a batch-mismatched latent.
With $\widetilde z_t$ obtained by permuting latent actions across samples, the
we impose the ranking constraint
\begin{align}
 \mathcal{L}_{\mathrm{dep}}=
 \Big[\mu+\|D_\psi(F_t,\widehat z_t)-\Delta F_t\|_2^2\nonumber\\
 \qquad-\|D_\psi(F_t,\widetilde z_t)-\Delta F_t\|_2^2\Big]_+ .
 \label{eq:dependency}
\end{align}
Thus, the matched code is required to explain the observed feature motion
better than an unrelated code, directly testing the claimed world--action
coupling.

\noindent\textbf{Flow-matching action policy.}
The action head is a conditional transformer with flow matching. It consumes
$F_t$, $\widehat F_{t+1}$, the full VLM hidden sequence, proprioception, and an
embodiment identifier. For robot samples, it is trained to match the vector
field of a continuous interpolation between noise and the demonstrated action
chunk. For $\epsilon\sim\mathcal{N}(0,I)$ and $\tau\in[0,1]$,
\begin{align}
    a_\tau&=(1-\tau)\epsilon+\tau a,\nonumber\\
    u(a,\epsilon,\tau)&=a-\epsilon,\nonumber\\
    \mathcal{L}_{\mathrm{flow}}
    &=\mathbb{E}_{\tau,\epsilon}
      \left\|v_\theta(a_\tau,c)-u(a,\epsilon,\tau)\right\|_2^2 .
    \label{eq:flow}
\end{align}
where $c=(F_t,\widehat F_{t+1},h_t^{\mathrm{vlm}},s_t)$ and $a_\tau$ is the
noised action at flow time $\tau$. At inference, numerical integration of the
learned vector field produces an action chunk of the requested horizon.

The action loss is evaluated only on valid, non-padded action tokens and only
for samples carrying robot actions. The flow head embeds the action time grid,
proprioception, and embodiment identifier, and receives the concatenation of
the current feature, the reconstructed future feature, and the VLM context.
During training, the visual condition can mix predicted and ground-truth future
features:
\begin{align}
 F_{t+1}^{\mathrm{cond}}&=r_t\widehat F_{t+1}+(1-r_t)F_{t+1},\nonumber\\
 c_t&=(F_t,F_{t+1}^{\mathrm{cond}},h_t^{\mathrm{vlm}},s_t),
 \label{eq:flow_mixing}
\end{align}
where $r_t$ is a Bernoulli mask whose probability of selecting the prediction
is increased during training. At inference $r_t=1$, so the action head sees the
same predicted-future condition that it receives after deployment.

The future feature is an internal predictive condition, not an additional
observation. Consequently, the policy must use the latent action to form a
useful future state if it is to exploit this pathway at deployment.

\subsection*{\textbf{C. End-to-end training architecture}}
\label{sec:training}

Figure~\ref{fig:overview} is also the training computation graph. For every
clip, the first observation and instruction enter the online $X$-VLM once,
yielding both $\widehat z_t$ and the context $h_t^{\mathrm{vlm}}$. The predicted
code is passed to delta-lwm, whose future feature is concatenated with the
current feature and VLM context at the flow head. The robot-action loss thus
has a differentiable route through the flow head, predicted future feature,
world decoder, latent-action predictor, and online VLM. The temporal and world
losses supervise the same intermediate quantities; all active losses are
accumulated before one optimizer update. The EMA branch only provides the
stop-gradient target for temporal alignment and is updated after that step.

\begin{table}[!t]
\begin{minipage}[t]{\columnwidth}
\centering
\caption{\normalfont\textbf{Performance comparison on standard LIBERO.}
Success rates (\%) are reported across four suites; Avg. is their unweighted mean.}
\label{tab:libero}
\small
\setlength{\tabcolsep}{2pt}
\begin{tabular}{@{}lrrrrr@{}}
\toprule
Method & Spatial & Object & Goal & Long & Avg. \\
\midrule
$\pi_0$~\cite{black2025pi0} & 98.0 & 96.8 & 94.4 & 88.4 & 94.4 \\
$\pi_0$-FAST~\cite{pertsch2025fast} & 96.4 & 96.8 & 88.6 & 60.2 & 85.5 \\
$\pi_{0.5}$~\cite{intelligence2025pi05} & \textbf{98.8} & 98.2 & \textbf{98.0} & 92.4 & 96.9 \\
WorldVLA~\cite{cen2025worldvla} & 87.6 & 96.2 & 83.4 & 60.0 & 81.8 \\
LAPA~\cite{ye2025lapa} & 55.4 & 58.8 & 74.6 & 73.8 & 65.7 \\
Mantis~\cite{yang2025mantis} & \textbf{98.8} & 99.2 & 94.4 & \textbf{94.2} & 96.7 \\
UniVLA~\cite{bu2025univla} & 96.5 & 96.8 & 95.6 & 92.0 & 95.2 \\
VLA-JEPA~\cite{sun2026vlajepa} & 94.8 & 99.6 & 95.8 & 94.0 & 96.1 \\
\midrule
TS-WAM & 97.2 & \textbf{99.8} & 97.6 & 93.4 & \textbf{97.0} \\
\bottomrule
\end{tabular}
\end{minipage}\par\vspace{8pt}
\begin{minipage}[t]{\columnwidth}
\centering
\caption{\normalfont\textbf{Performance comparison on SimplerEnv.}
Success rates (\%) are reported for four task categories; Avg. is their unweighted
mean. VM and VA denote visual matching and visual aggregation, respectively.}
\label{tab:simplerenv}
\small
\setlength{\tabcolsep}{3pt}
\begin{tabular}{@{}lrrrrr@{}}
\toprule
Method & \multicolumn{2}{c}{WindowX Robot} & \multicolumn{2}{c}{Google Robot} & Avg. \\
 & \multicolumn{2}{c}{(VM)} & \multicolumn{2}{c}{(VA)} & \\
\cmidrule(lr){2-3}\cmidrule(lr){4-5}
 & Spoon & Carrot & Pick & Move & \\
\midrule
RT-1-X~\cite{embodiment2023rtx} & 0.0 & 4.2 & 49.0 & 32.3 & 21.38 \\
Octo-Base~\cite{octo2024policy} & 15.8 & 12.5 & 0.6 & 3.1 & 8.00 \\
OpenVLA~\cite{kim2024openvla} & 4.2 & 0.0 & 60.8 & 67.7 & 33.18 \\
Magma~\cite{yang2025magma} & 37.5 & 31.0 & 68.8 & 65.7 & 50.75 \\
$\pi_0$~\cite{black2025pi0} & 29.1 & 0.0 & 75.2 & 63.7 & 42.00 \\
$\pi_0$-FAST~\cite{pertsch2025fast} & 29.1 & 21.9 & 77.6 & 68.2 & 49.20 \\
RoboVLMs~\cite{li2024robovlms} & 45.8 & 20.8 & 77.3 & 61.7 & 51.40 \\
TraceVLA~\cite{zheng2024tracevla} & 12.5 & 16.6 & 64.4 & 60.6 & 38.53 \\
\midrule
TS-WAM & \textbf{62.5} & \textbf{37.5} & \textbf{80.33} & \textbf{75.0} & \textbf{63.83} \\
\bottomrule
\end{tabular}
\end{minipage}
\end{table}

\begin{table*}[!t]
\centering
\caption{\normalfont\textbf{Zero-shot transfer results on LIBERO-Plus.}
Success rates (\%) are reported across seven perturbation categories and overall.}
\label{tab:liberoplus}
\small
\setlength{\tabcolsep}{4pt}
\begin{tabular}{@{}lrrrrrrrrr@{}}
\toprule
Method & Params & Camera & Robot & Language & Light & Background & Noise & Layout & Overall \\
\midrule
WorldVLA~\cite{cen2025worldvla} & 7B & 0.1 & 27.9 & 41.6 & 43.7 & 17.1 & 10.9 & 38.0 & 25.0 \\
Mantis~\cite{yang2025mantis} & 5.8B & 15.7 & 41.8 & 45.9 & 45.1 & 28.9 & 39.2 & 62.5 & 39.8 \\
UniVLA~\cite{bu2025univla} & 7B & 4.3 & 50.3 & 71.8 & 59.1 & 80.0 & 25.3 & 34.3 & 41.5 \\
$\pi_0$~\cite{black2025pi0} & 3.3B & 13.8 & 6.0 & 58.8 & 85.0 & 81.4 & \textbf{79.0} & 68.9 & 53.6 \\
VLA-JEPA~\cite{sun2026vlajepa} & 2B & 40.3 & 55.7 & 72.9 & 88.2 & 70.5 & 38.2 & 74.6 & 62.9 \\
Spatial Forcing~\cite{li2025spatialforcing} & 7B & 20.1 & 13.4 & 40.9 & 29.1 & 33.4 & 25.7 & 39.3 & 28.8 \\
Fast-WAM~\cite{yuan2026fastwam} & 6B & 16.4 & 44.5 & 68.9 & 78.2 & 53.7 & 37.7 & 60.7 & 51.4 \\
\midrule
TS-WAM & 1.2B & \textbf{50.5} & \textbf{60.1} & \textbf{77.5} & \textbf{94.4} & \textbf{91.5} & 76.1 & \textbf{79.6} & \textbf{75.7} \\
\bottomrule
\end{tabular}
\end{table*}

We train this graph with mixed robot and ego (action-free human) clips. A
sample-level validity mask determines which terminal heads contribute, while
the shared online VLM, latent predictor, and world decoder are executed for
both data sources. For a robot clip, action, temporal, and world terms are
active:
\begin{equation}
 \mathcal{L}_{\mathrm{robot}}=\mathcal{L}_{\mathrm{flow}}
 +\lambda_T\mathcal{L}_{\mathrm{temp}}
 +\lambda_W\mathcal{L}_{\mathrm{world}} .
 \label{eq:total}
\end{equation}
The coefficients balance shared temporal and dynamics supervision against
control. An ego clip uses the same forward path and current--future feature
targets, but has no robot state or action; its masked objective is
\begin{equation}
 \mathcal{L}_{\mathrm{ego}}=\lambda_T\mathcal{L}_{\mathrm{temp}}
 +\lambda_W\mathcal{L}_{\mathrm{world}} .
 \label{eq:ego}
\end{equation}

Robot and ego samples can be interleaved in one batch or in successive
micro-batches. Their masked losses are summed into one update of the shared
modules, so ego videos continually shape the representation used by robot
control instead of forming a discarded pretraining stage. At deployment the
graph is truncated at the online branch, delta-lwm, and flow head; the
future-aware EMA branch and training-only targets are absent.

\begin{figure}[t]
    \centering
    \includegraphics[width=\columnwidth]{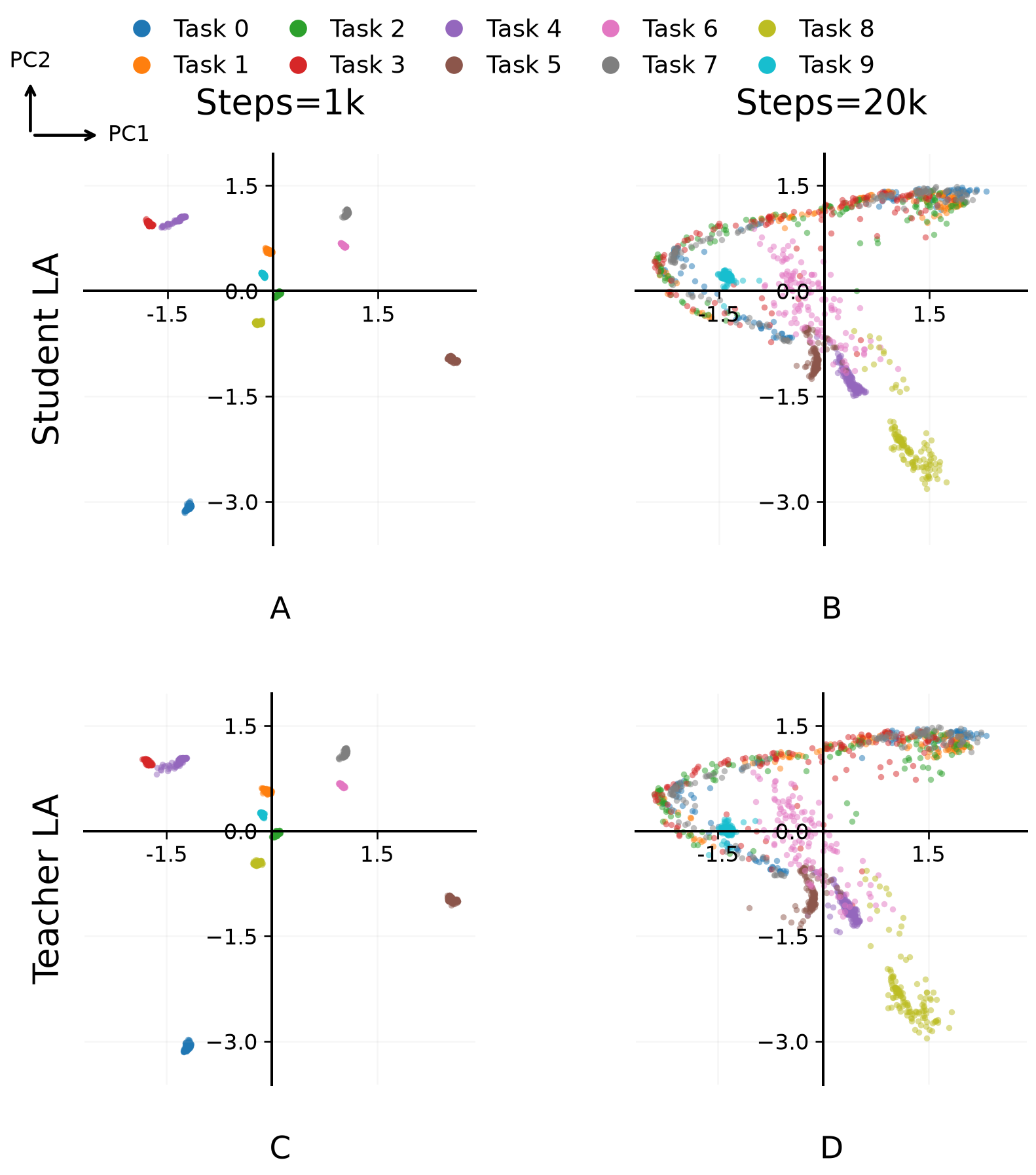}
    \caption{\textbf{Latent-action representation diversity.} Each point is a temporal
    window, projected onto two principal components; colors identify tasks.
    The upper and lower rows show student and teacher latent actions,
    respectively.}
    \label{fig:la_diversity}
\end{figure}

\section{Experiments}
\label{sec:experiments}

We evaluate whether learning temporal representations from data without action
labels improves manipulation and generalization across three simulation
platforms and two real-world robotic platforms.

\subsection{\textbf{Experimental Setup}}
\label{sec:experimental_setup}
\textbf{Benchmarks.} We evaluate TS-WAM on three simulation
benchmarks.
\begin{itemize}
\item \textbf{LIBERO.} We evaluate the Spatial, Object, Goal, and Long suites
to assess manipulation performance.
\item \textbf{LIBERO-Plus.} Building on LIBERO, we perform zero-shot transfer
and evaluate performance under seven categories of condition changes.
\item \textbf{SimplerEnv.} We evaluate four task categories from the Visual
Matching (VM) and Visual Aggregation (VA) suites on the WindowX Robot and
Google Robot.
\end{itemize}

\begin{figure}[!t]
    \centering
    \includegraphics[width=\columnwidth]{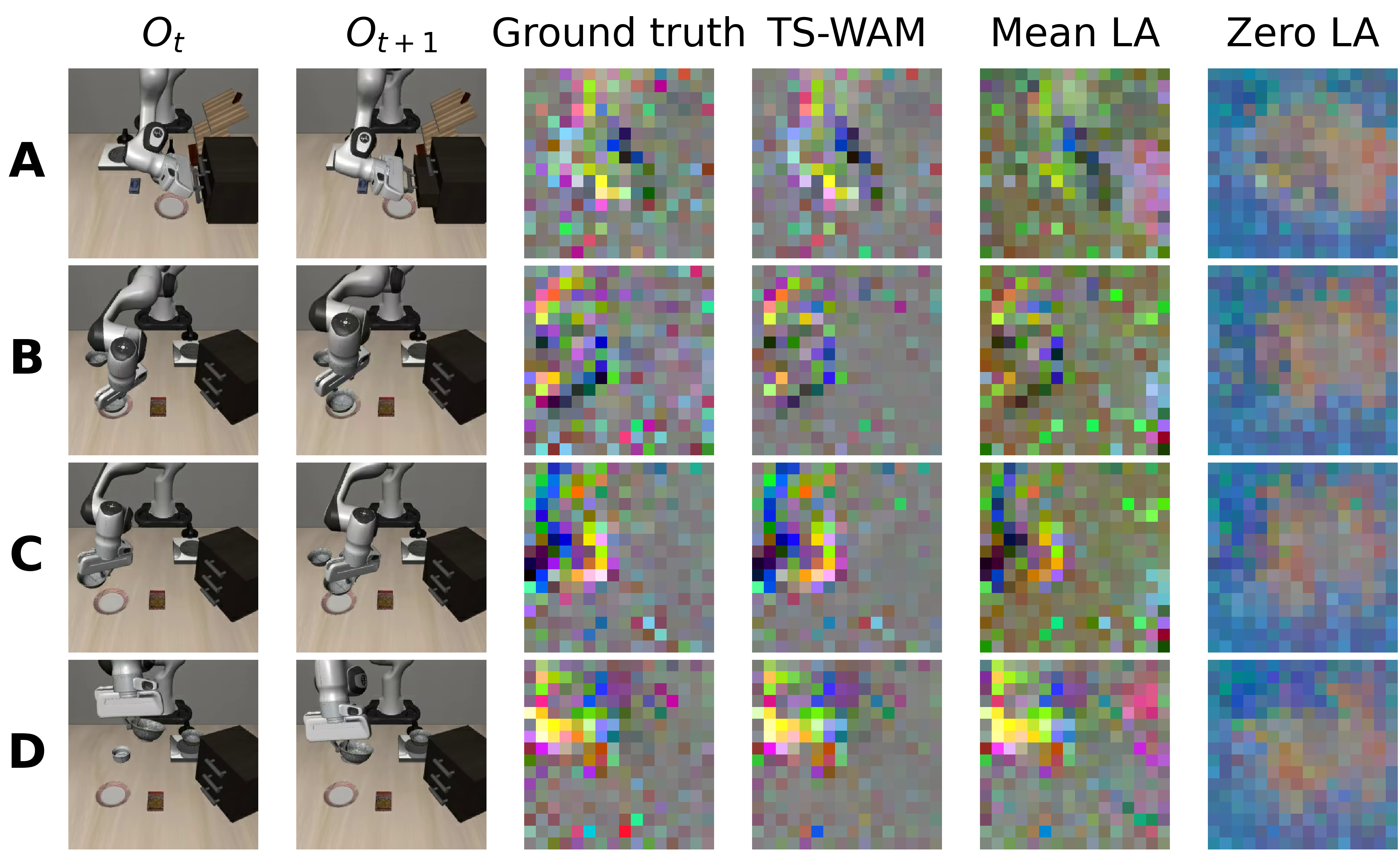}
    \caption{\textbf{Effect of latent-action conditioning on LWM predictions.}
    Rows A--D show four task examples. The first two columns contain the
    observed temporal pair; the remaining columns visualize the ground-truth
    feature change and predictions using the normal student LA (TS-WAM),
    a fixed mean LA, and a zero LA. All feature maps share the same
    GT-fitted PCA basis and RGB scale.}
    \label{fig:la_conditioning}
\end{figure}

For each benchmark, we select baselines from multiple method categories and
conduct extensive comparisons over repeated trials. We report task success
rates according to each benchmark's task-completion criterion.

\textbf{Real-world Setup.} We evaluate TS-WAM on two robotic platforms.
\begin{itemize}
\item \textbf{Franka Research 3.} We use a 7-DoF Franka
Research 3 single-arm robot with two RGB observation cameras. We evaluate bowl
stacking and a long-horizon drawer task that requires opening a drawer, placing
an object inside, and closing the drawer.
\item \textbf{RevoTron.} We use the RevoTron bimanual humanoid robot equipped
with Revo3 dexterous hands and three RGB observation cameras. The VLA policy jointly controls
both arms and the dexterous hands. We evaluate object sorting and the
long-horizon task of opening a drawer, placing an object inside, and closing
the drawer.
\end{itemize}

On both platforms, we compare TS-WAM with
$\pi_{0.5}$~\cite{intelligence2025pi05} using task success rates under the
original in-distribution (IND) conditions and three OOD settings:
\emph{OOD-Light conditions}, with an additional light switched on;
\emph{OOD-Objects Layout}, with a changed object arrangement; and
\emph{OOD-Background Textures}, with a black cloth covering the original
white tabletop.

\begin{table}[!t]
    \centering
    \caption{\normalfont\textbf{Real-world performance comparison under IND and OOD conditions.}
    Success rates (\%) are reported over 20 trials per task and condition;
    Overall is the mean across four tasks.
    OOD-LC, OOD-OL, and OOD-BT denote OOD-Light Conditions,
    OOD-Objects Layout, and OOD-Background Textures, respectively.}
    \label{tab:real_world}
    \small
    \setlength{\tabcolsep}{2pt}
    \renewcommand{\arraystretch}{1.0}
    \begin{tabular}{@{}lrrrrr@{}}
        \toprule
        Method & \multicolumn{2}{c}{Franka Research 3} & \multicolumn{2}{c}{RevoTron} & Overall \\
        \cmidrule(lr){2-3}\cmidrule(lr){4-5}
         & \shortstack{Stack\\Bowls} & \shortstack{Drawer-\\Long} & \shortstack{Item\\Sorting} & \shortstack{Drawer-\\Long} & \\
        \midrule
        $\pi_{0.5}$ & & & & & \\
        \hspace{2em}IND & 75 & 40 & 60 & 45 & 55.00 \\
        \hspace{2em}OOD-LC & 60 & 20 & 50 & 35 & 41.25 \\
        \hspace{2em}OOD-OL & 45 & 20 & 40 & 25 & 32.50 \\
        \hspace{2em}OOD-BT & 40 & 25 & 30 & 30 & 31.25 \\
        \midrule
        TS-WAM & & & & & \\
        \hspace{2em}IND & \textbf{85} & \textbf{65} & \textbf{80} & \textbf{60} & \textbf{72.50} \\
        \hspace{2em}OOD-LC & \textbf{80} & \textbf{55} & \textbf{75} & \textbf{60} & \textbf{67.50} \\
        \hspace{2em}OOD-OL & \textbf{80} & \textbf{45} & \textbf{70} & \textbf{50} & \textbf{61.25} \\
        \hspace{2em}OOD-BT & \textbf{70} & \textbf{45} & \textbf{70} & \textbf{55} & \textbf{60.00} \\
        \bottomrule
    \end{tabular}
\end{table}

\textbf{Implementation details.} We deploy TS-WAM on our in-house real-world
robot control platform with an inference frequency of 30\,Hz. Training runs on
a single machine with four NVIDIA A100 GPUs, while inference runs on a single
consumer-grade NVIDIA RTX 4090 GPU.

\subsection{\textbf{Simulation Results}}
\label{sec:simulation_results}

\textbf{The joint architecture learns informative latent-action representations.}
The PCA visualization in Fig.~\ref{fig:la_diversity} shows a compact latent
space early in training that becomes substantially richer by the end of
training, with variation preserved within individual tasks rather than collapse
to a task prototype. This temporal structure is learned jointly from robot and
action-free video clips and supplies the representation used by downstream
control modules.

\begin{figure}[!t]
\centering
\includegraphics[width=\columnwidth]{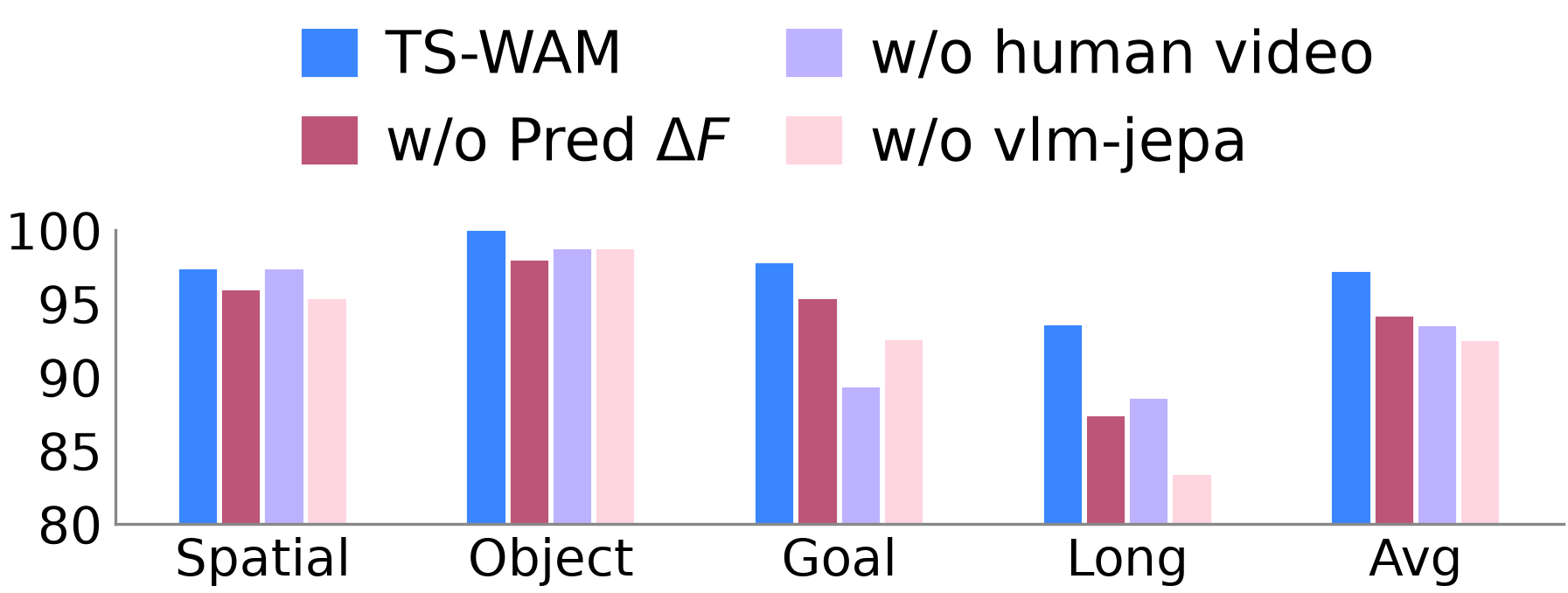}
\caption{\textbf{LIBERO ablation results.}
``w/o human video'' denotes training without ego data;
``w/o Pred delta F'' denotes an action expert that receives only semantic
features from the VLM, without future features from delta-lwm;
``w/o vlm-jepa'' denotes training with a single VLM without the JEPA architecture.}
\label{fig:libero_ablation}
\par\vspace{8pt}
\centering
\includegraphics[width=\columnwidth,trim={0 15bp 0 0},clip]{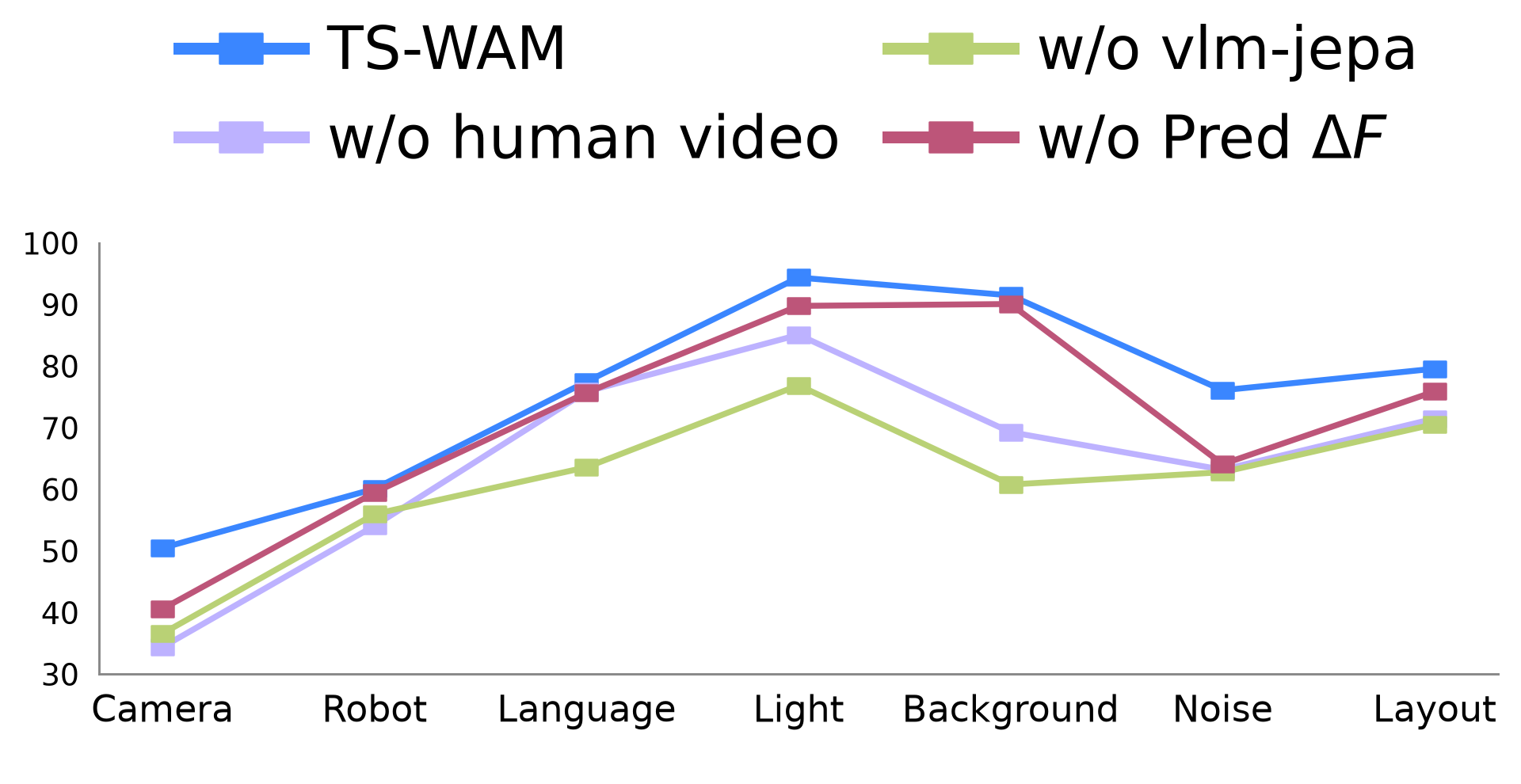}
\caption{\textbf{LIBERO-PLUS ablation results.}
The ablation settings are described in the caption of Fig.~\ref{fig:libero_ablation}.}
\label{fig:libero_plus_ablation}
\end{figure}

\textbf{Delta-lwm's prediction of future feature changes depends strongly on
latent-action guidance.} In Fig.~\ref{fig:la_conditioning}, predictions made
with the normal sample-specific latent action nearly match the ground-truth
feature-increment distribution across four LIBERO task classes. Replacing it
with the training-distribution mean or removing it entirely produces clearly
different PCA patterns. With current features and decoder parameters fixed,
this comparison isolates latent action as the source of the prediction. The
normal-LA prediction retains localized structures close to the observed change,
whereas mean-LA and zero-LA predictions are diffuse or misplaced.

\begin{figure*}[!t]
    \centering
    \includegraphics[width=\textwidth]{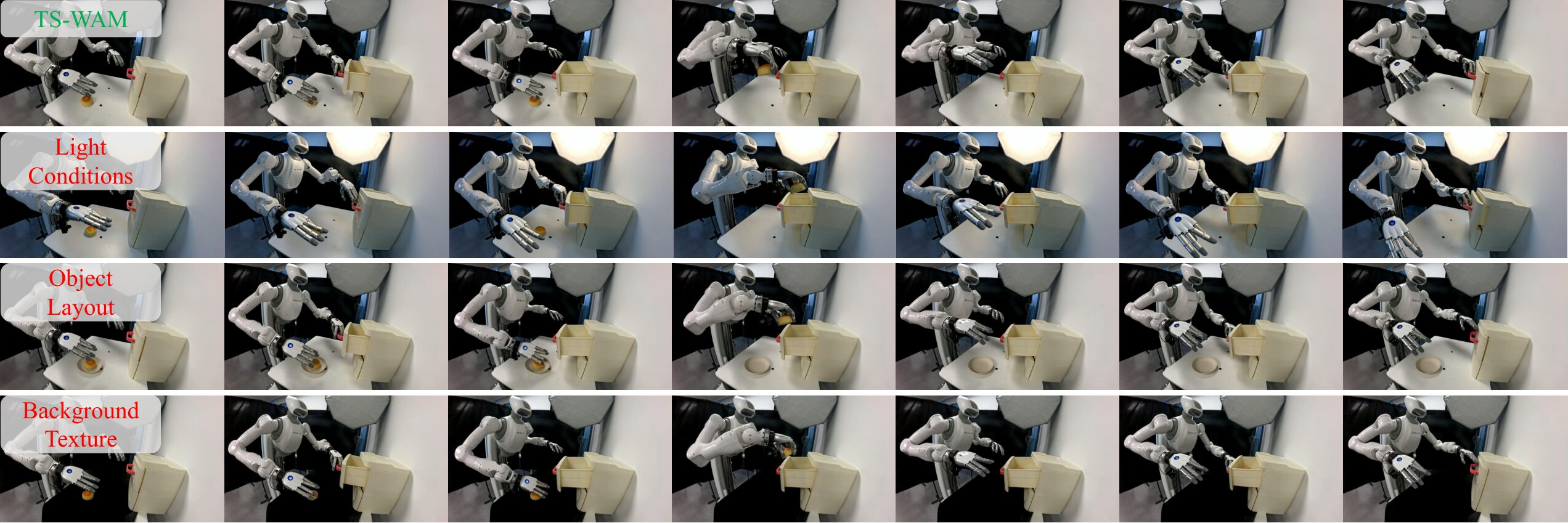}
    \caption{\textbf{TS-WAM executing Drawer-Long on RevoTron under IND and three
    OOD conditions.} Rows from top to bottom show the original IND scene,
    OOD-Light conditions (an additional light switched on), OOD-Objects
    Layout (bread placed on a plate instead of directly on the table), and
    OOD-Background Textures (the white tabletop covered with a black cloth).
    Time progresses from left to right. Each sequence shows successful
    completion of opening the drawer, placing the bread inside, and closing
    the drawer despite the visual changes.}
    \label{fig:real_world_ood}
\end{figure*}

\textbf{The end-to-end joint training architecture performs well across
simulation benchmarks.}

On LIBERO, TS-WAM reaches 97.0\% average success, compared with 96.9\% for
$\pi_{0.5}$ and 96.1\% for VLA-JEPA. It is best on the Object suite (99.8\%)
and remains competitive on Spatial (97.2\%), Goal (97.6\%), and Long (93.4\%),
showing that the joint model fits both short and long-horizon control
(Table~\ref{tab:libero}).

On SimplerEnv, TS-WAM averages 62.5\% and 37.5\% on the two Visual Matching
tasks (Spoon and Carrot), exceeding RoboVLMs (45.8\% and 20.8\%). On the two
Visual Aggregation tasks (Pick and Move), it reaches 80.33\% and 75.0\%,
improving over RoboVLMs' 77.3\% and 61.7\%. The gains across both WindowX
and Google robots indicate adaptation to distinct embodiments and evaluation
modes (Table~\ref{tab:simplerenv}).

On LIBERO-Plus, TS-WAM obtains 75.7\% overall, ahead of VLA-JEPA (62.9\%) and
$\pi_0$ (53.6\%). It leads VLA-JEPA on Camera (50.5\% vs. 40.3\%), Language
(77.5\% vs. 72.9\%), Light (94.4\% vs. 88.2\%), Background (91.5\% vs.
70.5\%), and Layout (79.6\% vs. 74.6\%), with the largest margin under
Background perturbations. These results support improved generalization and
robustness under distribution shifts (Table~\ref{tab:liberoplus}).

\subsection{\textbf{Real-World Results}}
\label{sec:real_world_results}

\textbf{The vlm-jepa--delta-lwm joint architecture transfers effectively to
real robots.} On the Franka Research 3 and RevoTron platforms, TS-WAM is
evaluated on four tasks spanning single-arm, bimanual, and dexterous
manipulation. It reaches 72.50\% mean success under the original
in-distribution (IND) scenes, compared with 55.00\% for the real-world
baseline $\pi_{0.5}$, and improves every platform--task pair in
Table~\ref{tab:real_world}. Thus, the joint latent-action and future-feature
pathway remains effective across embodiments and task horizons, including the
sequential drawer task.

\textbf{Joint learning from heterogeneous data improves real-world
generalization.} Relative to $\pi_{0.5}$, TS-WAM maintains higher success in
all 12 OOD platform--task and condition combinations. Its mean OOD success is
62.92\%, versus 35.00\% for $\pi_{0.5}$; the advantage is consistent across
lighting, object-layout, and background changes (67.50\%, 61.25\%, and
60.00\% versus 41.25\%, 32.50\%, and 31.25\%). Moreover, the IND-to-OOD
drop is only 9.58 percentage points for TS-WAM, compared with 20.00 points
for $\pi_{0.5}$. These results indicate that combining action-labeled robot
demonstrations with action-free video supervision strengthens robustness to
visual and spatial shifts while preserving the underlying manipulation goals.

\subsection{\textbf{Ablation Studies}}
\label{sec:ablations}

Figures~\ref{fig:libero_ablation} and~\ref{fig:libero_plus_ablation} report three
ablations of the full TS-WAM on standard LIBERO and LIBERO-Plus: (a) \emph{w/o human video} excludes additional human-video
data; (b) \emph{w/o vlm-jepa} removes the self-supervised JEPA architecture;
and (c) \emph{w/o Pred $\Delta F$} removes the cross-attention pathway that
introduces the predicted feature delta $\widehat{\Delta F}_t$ into the DiT
action head.

\textbf{Heterogeneous training with human video improves both basic control
and generalization.} On standard LIBERO, removing the additional human-video
data lowers the mean success from 97.00\% to 93.35\%, with the largest loss
on the Goal suite (8.4 points; Fig.~\ref{fig:libero_ablation}). This establishes
that action-free video contributes to core manipulation ability, even when
robot demonstrations are available. The same removal causes Overall success
on LIBERO-Plus to fall from 75.7\% to 64.8\%, showing that the benefit is
larger under the seven distribution-shift categories
(Fig.~\ref{fig:libero_plus_ablation}) and therefore extends to generalization.

\textbf{The vlm-jepa self-supervised architecture is the key to using the
heterogeneous data effectively.} Among the LIBERO ablations, removing
vlm-jepa causes the largest degradation: mean success drops to 92.35\% and
Long-suite success falls from 93.4\% to 83.2\%. Its effect is even clearer
on LIBERO-Plus, where the no-vlm-jepa model reaches only 61.0\% Overall,
below the human-video and Pred $\Delta F$ ablations (64.8\% and 70.8\%), and
is consistently weaker across the seven perturbation categories. Thus, the
JEPA interface is central to converting action-free temporal supervision into
useful representations for robot control and robust transfer.

\textbf{Predicted feature-delta conditioning enables the action expert to use
future information.} Removing the Pred $\Delta F$ pathway reduces LIBERO mean
success by 3.0 points, including a 6.2-point loss on Long, and lowers
LIBERO-Plus Overall success by 4.9 points to 70.8\%. The consistent drop in
both standard control and shifted-condition evaluation indicates that feeding
the predicted future feature to the action expert provides a positive,
task-relevant signal rather than redundant visual context.

\section{LIMITATIONS}
\label{sec:limitations}

TS-WAM introduces additional training costs. The EMA teacher branch modestly
increases GPU memory consumption during training. In addition, conditioning
the DiT action expert on predicted future features slows its convergence,
leading to a longer overall training time. These limitations motivate the use
of data partitioning to reduce GPU memory consumption and shared attention
to improve the convergence efficiency of the action expert.

\section{CONCLUSION}
\label{sec:conclusion}

We presented TS-WAM, a lightweight Teacher-Student World-Action Model for
vision-language robot control. Through self-supervised temporal learning and
explicit future prediction, TS-WAM learns from heterogeneous robot
demonstrations and action-free human videos within a unified end-to-end
training framework. Experiments on LIBERO, LIBERO-Plus, and SimplerEnv,
together with real-world evaluations on Franka Research 3 and RevoTron,
demonstrate strong manipulation performance and improved generalization
and robustness.

\bibliographystyle{IEEEtran}
\bibliography{references}

\end{document}